\pdfoutput=1

\documentclass[11pt]{article}

\usepackage[preprint]{acl}

\usepackage{times}
\usepackage{latexsym}

\usepackage[T1]{fontenc}

\usepackage[utf8]{inputenc}

\usepackage{microtype}

\usepackage{inconsolata}

\usepackage{graphicx}

\usepackage{times}
\usepackage{soul}
\usepackage{url}
\usepackage[utf8]{inputenc}
\usepackage{amsmath}
\usepackage{amsthm}
\usepackage{booktabs}
\usepackage[switch]{lineno}
\usepackage{makecell}
\usepackage{caption}
\usepackage{subfigure}
\usepackage{booktabs}
\usepackage{algorithm}
\usepackage{algpseudocode}

\usepackage{subcaption}
\usepackage{multirow}
\usepackage{xcolor}

\title{S$^4$C: Speculative Sampling with Syntactic and Semantic Coherence for \\ Efficient Inference of Large Language Models}

\author{
 \textbf{Tao He\textsuperscript{1,2}\thanks{Equal contribution. \href{mailto:kevin.92.he@gmail.com}{kevin.92.he@gmail.com}, \href{mailto:mc35324@connect.um.edu.mo}{mc35324@connect.um.edu.mo}}},
 \textbf{Guang Huang\textsuperscript{3}\footnotemark[1]},
 \textbf{Xu Yang\textsuperscript{1}},
 \textbf{Tianshi Xu\textsuperscript{1}},
\\
 \textbf{Sicheng Zhao\textsuperscript{4}},
 \textbf{Guiguang Ding\textsuperscript{4}},
 \textbf{Pengyang Wang\textsuperscript{3}},
 \textbf{Feng Tian\textsuperscript{1}\thanks{Corresponding author.}}
\\
\\
 \textsuperscript{1}GRG Banking Equipment Co., Ltd,
 \textsuperscript{2}South China University of Technology, \\
 \textsuperscript{3}University of Macau,
 \textsuperscript{4}Tsinghua University
 \\
}

\begin{document}
\maketitle
\begin{abstract}
Large language models (LLMs) exhibit remarkable reasoning capabilities across diverse downstream tasks. However, their autoregressive nature leads to substantial inference latency, posing challenges for real-time applications. Speculative sampling mitigates this issue by introducing a drafting phase followed by a parallel validation phase, enabling faster token generation and verification. Existing approaches, however, overlook the inherent coherence in text generation, limiting their efficiency. 
To address this gap, we propose a \textit{Speculative Sampling with Syntactic and Semantic Coherence}  (S$^4$C) framework, which extends speculative sampling by leveraging multi-head drafting for rapid token generation and a continuous verification tree for efficient candidate validation and feature reuse. 
Experimental results demonstrate that S$^4$C surpasses baseline methods across mainstream tasks, offering enhanced efficiency, parallelism, and the ability to generate more valid tokens with fewer computational resources. On Spec-bench benchmarks, S$^4$C achieves an acceleration ratio of 2.26x-2.60x, outperforming state-of-the-art methods.
\end{abstract}

\section{Introduction}
The rapid advancement of large language models (LLMs) has significantly improved performance across various downstream tasks and made real-time human-computer interaction an essential part of daily life, offering substantial convenience to users~\cite{achiam2023gpt,touvron2023llama,touvron2023llama2,chiang2023vicuna,jiang2023mistral}. However, the autoregressive transformer decoder architecture adopted by LLMs introduces substantial inference latency, limiting their deployment in real-time applications. As the generated sequence length and model size increase, the token-by-token serial generation process leads to escalating delays.

To address this challenge, speculative sampling~\cite{stern2018blockwise,leviathan2023fast,chen2023accelerating,xia2023speculative} has been proposed. This approach divides the inference process into a low-cost drafting phase and a parallel validation phase, allowing multiple tokens to be verified within a single LLM forward pass. By generating multiple tokens per pass, speculative sampling significantly accelerates text generation. More importantly, the validation phase ensures that the generated text aligns with the original LLM's distribution, preserving output integrity.

\begin{figure}[!t]
	\centering
    \includegraphics[width=0.45\textwidth]{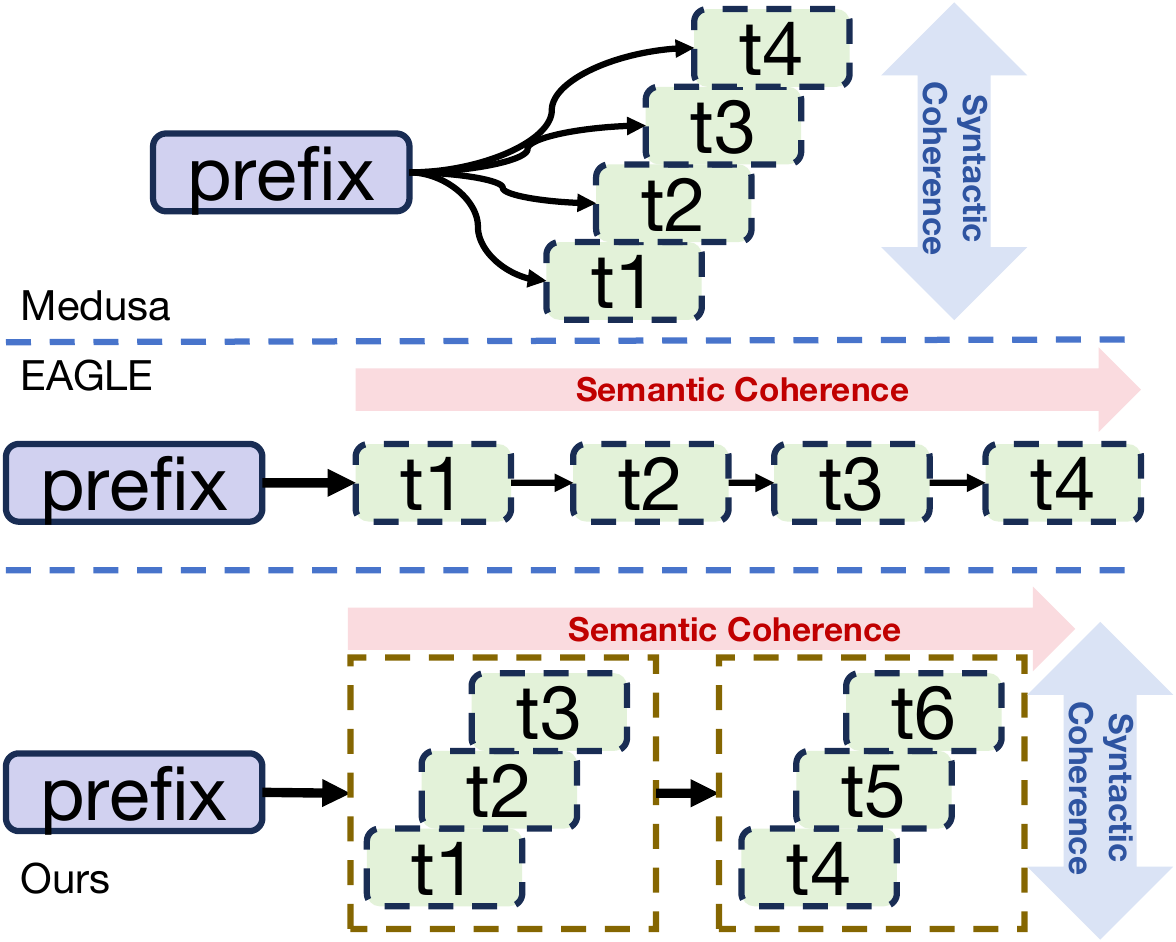}
    \caption{Comparison of different drafting processes. Medusa (top) generates four tokens in parallel based on a prefix. EAGLE (middle) follows an autoregressive approach, generating one token at a time. Our method (bottom) employs multi-round autoregression, generating multiple tokens at different positions in each round.}
	\label{fig1-2}
\end{figure}

The effectiveness of speculative sampling depends on selecting an appropriate draft model that approximates the original LLM while reducing latency. Typically, this is achieved by a smaller-parameter model from the same series~\cite{leviathan2023fast,chen2023accelerating}, such as TinyLLaMA~\cite{zhang2024tinyllama}, which serves as the draft model for the 7B and 13B versions of LLaMA2~\cite{touvron2023llama2}. While leveraging smaller draft models offers advantages, it also requires additional effort to train or select a model that closely aligns with the target LLM, presenting challenges in scalability and compatibility.
To overcome these challenges, recent studies have exploited the target LLM itself for drafting. As shown in Figure~\ref{fig1-2}, Medusa~\cite{cai2024medusa} generates multiple drafts in parallel based on prefixes, whereas EAGLE~\cite{li2024eagle} employs a sequential autoregressive drafting process to enhance precision.

To better analyze the advantages and disadvantages of these methods, we define two concepts: \textit{Syntactic Coherence} and \textit{Semantic Coherence}.
\begin{itemize}
    \item \textbf{Syntactic Coherence} refers to the adherence to fixed language collocations and structures that are commonly accepted in a language, without heavily relying on prior context.
    \item \textbf{Semantic Coherence} ensures that the meaning of the generated text aligns logically with the preceding context, maintaining overall textual consistency.
\end{itemize}
It can be observed that methods such as Medusa~\cite{cai2024medusa} eliminate the dependency between heads, thereby accelerating the generation of drafts. However, these methods primarily focus on modeling \textit{Syntactic Coherence} while neglecting \textit{Semantic Coherence}. On the other hand, methods like Eagle~\cite{li2024eagle}, which use an autoregressive approach, are more suitable for modeling \textit{Semantic Coherence}. Using such methods to model \textit{Syntactic Coherence} introduces unnecessary computational overhead.




In this work, we propose a \textit{\textbf{S}peculative \textbf{S}ampling framework with \textbf{S}yntactic and \textbf{S}emantic \textbf{C}oherence} for efficient inference of large language models, termed as \textbf{S$^4$C}. It consists of two primary components: the draft model and validation tree. 
By adopting a continuous multi-head structure, S$^4$C can efficiently generate syntactically coherent tokens and ensure semantic coherence between multiple fragments. On this basis, S$^4$C constructed a continuous validation tree, further enhancing the coherence of candidate paths by selecting the candidate token with the highest probability. Compared to existing methods, S$^4$C has a simpler structure, higher efficiency, and stronger parallelization capability. It can generate more effective tokens without increasing computational overhead.
The main contributions of this paper are as follows:
\begin{itemize}
    \item We identify the critical role of coherence in speculative decoding and propose the S$^4$C framework to address this challenge.
    \item We introduce a simple yet efficient multi-head continuous draft model that rapidly generates coherent token sequences while maintaining generation quality.
    \item We design a continuous verification tree that expands the candidate set with minimal computational cost, making it adaptable to various LLM architectures.
    \item Extensive experiments demonstrate that S$^4$C achieves superior performance compared to existing baseline methods.
\end{itemize}

The remainder of this paper is structured as follows: Section~\ref{sec:pre} provides preliminary background. Section~\ref{sec:method} describes the core modules of S$^4$C. Section~\ref{sec:exp} discusses experimental results. Related work and conclusions are presented in Sections~\ref{sec:related_work} and ~\ref{sec:conclusion}, respectively.

\section{Preliminaries}
\label{sec:pre}

\paragraph{Notations}
In this paper, we define the key terms as follows. The term ``target LLM'' refers to the large language model responsible for verifying tokens, denoted by $M_p$. The ``draft model'' is the model used to generate draft tokens, represented by $M_q$. The term ``feature'' denotes the output of the penultimate layer of the LLM, corresponding to the hidden state before the final prediction layer. A single token is represented by $t$, with its embedding denoted as $e$, its features as $f$, and its probability distribution as $p$.

\paragraph{Speculative Sampling}
Speculative sampling is a two-stage process comprising an initial drafting phase followed by a verification phase. In the drafting phase, a smaller draft model generates a set of $\gamma$ candidate tokens, denoted as $\hat{T}_{j+1:j+\gamma}$, along with their probability distributions $q$. The verification phase consists of a single forward pass through the target LLM to obtain the corresponding probabilities $p$. Each drafted token $\hat{t}_{j+i}$ is accepted with a probability of $\min(1, \frac{p}{q})$. If a token is rejected, subsequent tokens are discarded and resampled from $ \text{norm}(\max(0, p - q))$, as described in \cite{leviathan2023fast}.

Speculative sampling significantly reduces inference latency by processing multiple tokens in parallel while ensuring that the output remains consistent with the distribution of the target LLM. This method effectively balances efficiency and accuracy, making it a promising solution for accelerating autoregressive text generation.


\section{Methodologies}
\label{sec:method}
\begin{figure*}[th]
	\centering
    \includegraphics[width=0.85\textwidth]{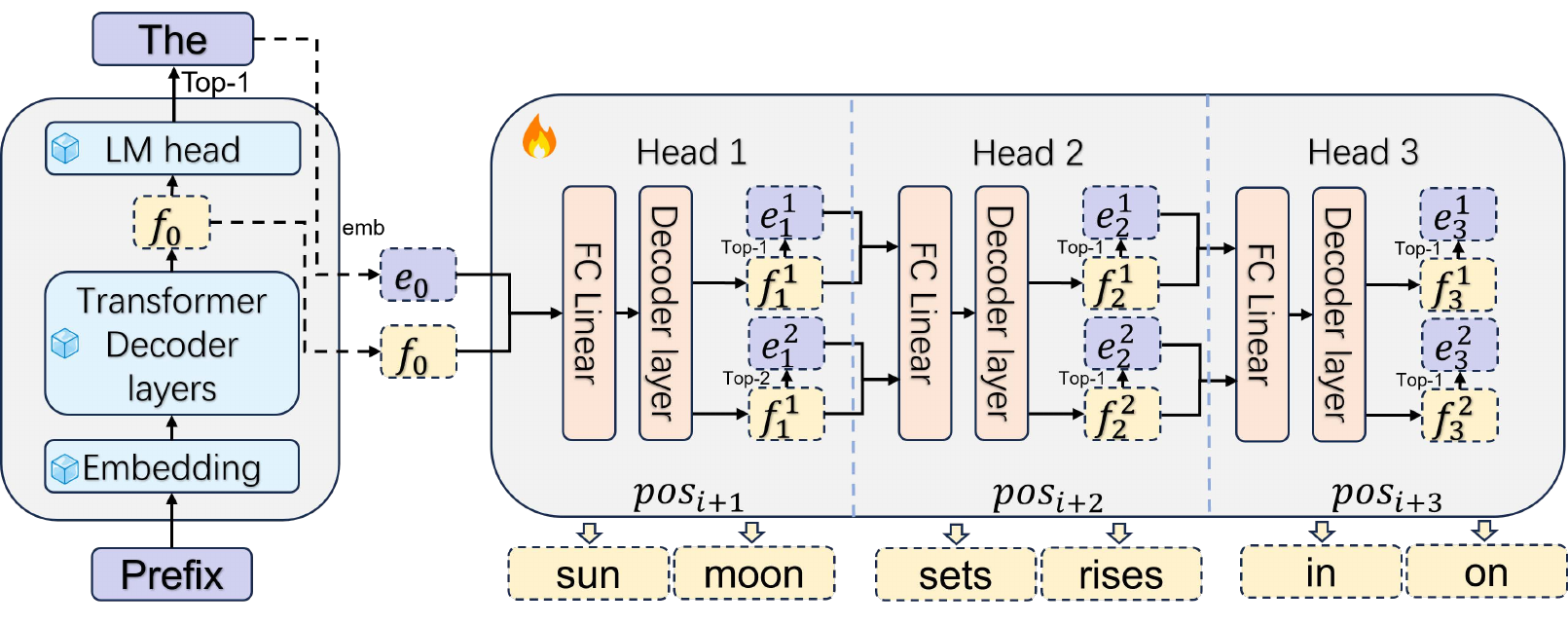} 
	\caption{Multi-head auto-regressive drafting architecture. The left side represents the target model with frozen parameters, while the right side illustrates the draft model with three heads.}
	\label{FIG:2}
\end{figure*}

Our proposed approach, \textit{\textbf{S}peculative \textbf{S}ampling framework with \textbf{S}yntactic and \textbf{S}emantic \textbf{C}oherence} (S$^4$C), follows the speculative sampling framework and consists of two main components: the drafting stage and the verification stage.

\subsection{Multi-head Auto-regressive Drafting}
\subsubsection{Architecture}
As illustrated in Figure~\ref{FIG:2}, the target model (left side) has frozen parameters. Initially, an initial forward pass is performed through the target model to process the input prefix, similar to standard Large Language Model (LLM) operations, which can be formalized as:

\begin{equation}
f_0 = \text{Decoder\_layers}(\text{Embedding}(t_{\text{pre}})),
\end{equation}
\begin{equation}
t_0 = \text{LM\_head}(f_0).
\end{equation}

The embedding of the prefix token sequence \( t_{\text{pre}} \) is processed by the decoder, producing an intermediate representation \( f_0 \). The language model head is subsequently applied to this representation to generate the initial token \( t_0 \), which establishes the foundational context for further token generation.

S$^4$C leverages the intermediate feature $f_0$ from the target model, bypassing the LM\_head layer. The generated token \( t_0 \) is then transformed into its embedding representation $e_0$ via the target model's embedding layer, serving as the input to the draft model. The draft model comprises three identical heads, defined as:

\begin{equation}
h_{i+1} = \text{Linear}(\text{concat}[e_{i}, f_{i}]), 
\end{equation}
\begin{equation}
f_{i+1} = \text{Decoder\_layers}(h_{i+1}).
\end{equation}

In this process, the input embeddings $e_0$ and features $f_0$ are concatenated and passed through a linear transformation to produce $h_{i+1}$, ensuring dimensional consistency with the original $f_0$. Subsequently, decoder layers generate feature vectors $f_{i:i+k}$, and token embeddings $e_{i:i+k}$ are obtained using the LM head and embedding layer of the target model.

The generated token \( t_{i+1} \) and its corresponding embedding \( e_{i+1} \) are obtained using the following equations:

\begin{equation}
t_{i+1} = \text{Argmax}(\text{LM\_head}(f_{i+1})),
\end{equation}
\begin{equation}
e_{i+1} = \text{Embedding}(t_{i+1}).
\end{equation}
In the first draft head, since only one feature from the target model is available, this feature is reused, selecting the top-2 candidates as different inputs. In the subsequent heads, the most probable word from $f_i$ is directly used to obtain $e_i$ through the embedding layer, ensuring efficient and coherent token generation.
In this way, each head generates multiple tokens simultaneously within itself, without dependencies between them, which is used to model \textit{syntactic coherence}. Between heads, an autoregressive approach is used, where the input of each head depends on the output of the previous head, which is used to model \textit{semantic coherence}.

\subsubsection{Training}
Our training process follows the same next-token regression task as used in Medusa~\cite{cai2024medusa}, Hydra~\cite{ankner2024hydra}, and EAGLE~\cite{li2024eagle}, integrating three loss components. The first component, $loss_{lm}$, represents the cross-entropy loss (standard LLM training loss) between predicted tokens and ground-truth labels, defined as:

\begin{equation}
loss_{lm} = -\sum_{i=1}^{n} {y_{i} \log{\hat{y_i}}}
\end{equation}
where $\hat{y_i}$ is the predicted token probability distribution from the LM head, and $y_i$ is the ground-truth token ID.

The other two loss components, $loss_{teacher}$ and $loss_{smooth}$, utilize cross-entropy and Smooth L1 loss, respectively, to measure the discrepancy between features generated by each draft model head and the target model. These are defined as:

\begin{equation}
loss_{teacher} = -\sum_{i=1}^{n} {p_i \log{q_i}}
\end{equation}

\begin{equation}
loss_{smooth} = 
\begin{cases} 
0.5(q_i - p_i)^2 & \text{if } |q_i - p_i| < 1 \\
|q_i - p_i| - 0.5 & \text{otherwise}
\end{cases}
\end{equation}
Here, $q_i$ represents the output of the draft model, while $p_i$ denotes the corresponding output of the target model.

The final loss function is a weighted sum of the three components, with weights set to $w_1 = 0.1$, $w_2 = 1.0$, and $w_3 = 0.1$, defined as:

\begin{equation}
loss = w_1 \cdot loss_{lm} + w_2 \cdot loss_{teacher} + \\ 
w_3 \cdot loss_{smooth}
\end{equation}

The tri-component loss function is designed to enhance the predictive capabilities of the draft model. The primary loss, $loss_{lm}$, ensures that the draft model accurately predicts the next token, aligning with conventional LLM objectives.

\begin{figure*}[h]
	\centering
    \includegraphics[width=0.9\textwidth]{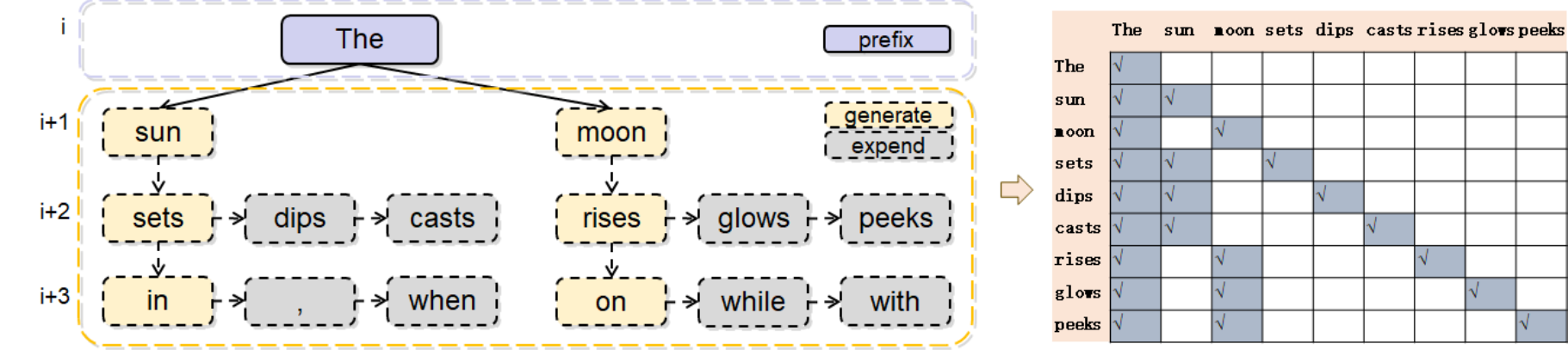} 
	\caption{Continuous verification tree architecture (left) and tree mask matrix (right).}
	\label{FIG:3}
\end{figure*}

In addition, the complementary losses $loss_{teacher}$ and $loss_{smooth}$ minimize the discrepancy between the draft and target models, facilitating knowledge transfer. This combined approach allows the draft model to effectively capture the predictive characteristics of the target model, enhancing its learning and generalization capabilities.

\subsection{Continuous Verification Tree}
In traditional rejection sampling, proposed tokens are structured as linear sequences, where the rejection of a token necessitates the discarding of all subsequent tokens. This approach limits proposal flexibility and reduces the acceptance rate of valid tokens. 

To overcome this limitation, we adopt a tree-based proposal structure inspired by previous studies~\cite{miao2024specinfer,cai2024medusa,li2024eagle}. This structure enables the exploration of alternative branches when a token is rejected, allowing for simultaneous validation of multiple candidate drafts at the same position. Consequently, it significantly increases the length of accepted drafts and enhances both efficiency and flexibility.

As illustrated in Figure~\ref{FIG:3}, our verification tree begins with the prefix ``The'' as the root node and generates multiple drafts vertically first (yellow tokens). Once the token set is validated, it extends horizontally (gray tokens). Vertically generated tokens at different positions correspond to the top-1 probability choices, leveraging efficient parallelization. For instance, in the figure, ``sets'' is the highest-probability token following ``sun'', while ``rises'' is the highest-probability token following ``moon''.

The horizontal expansion provides alternative token choices at the same position, offering top-k alternatives when the top-1 token is incorrect, thereby improving the overall acceptance rate. In Figure~\ref{FIG:3}, ``dips'' and ``casts'' represent top-3 probable tokens following ``sets'', while ``glows'' and ``peeks'' follow ``rises''.

Once the verification tree is constructed, a tree mask is applied to determine the longest accepted path through the entire tree.

After establishing the candidate set, we employ speculative sampling~\cite{leviathan2023fast} to verify token acceptance, formulated as follows:

\begin{equation}
r < \min\left(1,\frac{q(x)}{p(x)}\right), \quad r \sim U[0,1]
\end{equation}

where $r$ is a random variable drawn from the uniform distribution $U[0,1]$, and $q(x)$ and $p(x)$ denote the probabilities of token $x$ from the target and draft models, respectively. If $q(x) \geq p(x)$, the token $x$ is accepted. Otherwise, it is rejected with a probability of $1 - \frac{q(x)}{p(x)}$. 

To further ensure consistency with the target LLM's output distribution, the correction strategy~\cite{leviathan2023fast,miao2024specinfer} resamples tokens at fork positions using an adjusted distribution:

\begin{equation}
x_{t+c} \sim \text{norm}(\max(q_c, p_c))
\end{equation}

This process guarantees alignment with the target LLM's distribution while documenting accepted tokens and their corresponding features for subsequent drafting phases.

\section{Experiment}
\label{sec:exp}
In this section, we conduct comprehensive experiments to evaluate the performance of S$^4$C, addressing the following key research questions:

\noindent \textbf{RQ1:} How does S$^4$C improve inference speed compared to existing methods?  
This question evaluates S$^4$C's efficiency in reducing latency, a critical factor for real-world applications.

\noindent \textbf{RQ2:} What is the trade-off between additional space consumption and the achieved speedup ratio?  
Understanding this balance is essential for deploying S$^4$C in resource-constrained environments.

\noindent \textbf{RQ3:} What are the benefits of S$^4$C's multi-head structure in enhancing performance?  
This inquiry quantifies the impact of multi-head processing on model efficiency and versatility.

\noindent \textbf{RQ4:} How does the continuous validation tree contribute to overall performance?  
Evaluating the validation tree's role helps assess its impact on inference accuracy and efficiency.

\noindent \textbf{RQ5:} How robust is S$^4$C across different temperature settings during sampling?  
This analysis determines S$^4$C's stability under varying operational conditions.

\noindent \textbf{RQ6:} In which application scenarios does S$^4$C provide the most significant acceleration?  
Identifying key use cases helps clarify S$^4$C's practical benefits.

These experiments provide an in-depth evaluation of S$^4$C's effectiveness and establish benchmarks for its performance in accelerating large language model inference.

\subsection{Experiment Settings}
\subsubsection{Datasets and Models} 
We adopt the same experimental setup as Spec-Bench~\cite{xia2024unlocking}, which includes six subtasks: multi-round conversations, translation, summarization, question answering, mathematical reasoning, and retrieval-augmented generation. These correspond to the datasets MT-Bench~\cite{zheng2023judging}, WMT14 DE-EN, CNN/Daily Mail~\cite{nallapati2016abstractive}, Natural Questions~\cite{kwiatkowski2019natural}, GSM8K~\cite{cobbe2021training}, and DPR~\cite{karpukhin2020dense}, respectively.

We use Vicuna-v1.3~\cite{chiang2023vicuna} as the base model in three parameter sizes: 7B, 13B, and 33B. S$^4$C is evaluated against state-of-the-art models listed in the Spec-Bench leaderboard, including Eagle~\cite{li2024eagle}, Hydra~\cite{ankner2024hydra}, Medusa~\cite{cai2024medusa}, PLD~\cite{saxena2023prompt}, SPS~\cite{leviathan2023fast}, REST~\cite{he2023rest}, and Lookahead~\cite{fu2024break}.

\subsubsection{Evaluation and Environment}
To quantitatively assess the performance of inference acceleration techniques, we use the acceleration ratio as the primary metric, which measures speedup achieved during testing. Additionally, the average acceptance length is used to quantify the mean number of tokens accepted by the draft model, providing insights into token generation efficiency.

Experiments were conducted on two hardware setups. For the 7B model, experiments ran on a single NVIDIA A100 GPU with 40GB memory and 48 CPU cores. For the 13B and 33B models, we used four NVIDIA GeForce RTX 4090 GPUs (24GB each) and 96 CPU cores.

The software environment included PyTorch 2.4.0 with CUDA 12.6. To ensure consistency and isolate the effects of different methods, all experiments employed greedy decoding, FP16 precision, and a batch size of one. This standardized configuration ensures that variations in acceleration ratios and acceptance lengths are solely attributable to the inference acceleration methods under evaluation.

\subsection{Effectiveness (RQ1)}
\begin{table*}[!thbp]
  \centering

  \scalebox{0.67}{
    \begin{tabular}{l|l|cccccc|cc}
    \toprule
    Size & Models & \makecell{Multi-turn \\ Conversation} ↑ & \makecell{Trans- \\ lation} ↑ & \makecell{Summari \\ zation} ↑ & \makecell{Question \\ Answering}↑ & \makecell{Mathematical \\ Reasoning} ↑ & \makecell{Retrieval- \\ aug. \\ Generation}↑ & \makecell{Mean \\ Accepted \\ Tokens} ↑ & Overall ↑ \\
    \midrule
    \multirow{8}*{\textbf{7b}} 
    & EAGLE~\cite{li2024eagle} & 2.44x & 1.75x & 2.11x & 1.96x & 2.50x & 1.94x & 3.57 & 2.12x \\
    & Hydra~\cite{ankner2024hydra} & 2.52x & 1.89x & 1.81x & 1.91x & 2.40x & 1.76x & 3.25 & 2.06x \\
    & Medusa~\cite{cai2024medusa} & 2.04x & 1.65x & 1.58x & 1.65x & 1.97x & 1.49x & 2.32 & 1.73x \\
    & PLD~\cite{saxena2023prompt} & 1.64x & 1.04x & \textbf{2.43x} & 1.14x & 1.61x & 1.71x & 1.73 & 1.59x \\
    & SPS~\cite{leviathan2023fast} & 1.66x & 1.13x & 1.62x & 1.49x & 1.47x & 1.55x & 2.28 & 1.49x \\
    & REST~\cite{he2023rest} & 1.63x & 1.31x & 1.36x & 1.66x & 1.21x & 1.73x & 1.82 & 1.48x \\
    & Lookahead~\cite{fu2024break} & 1.40x & 1.14x & 1.19x & 1.24x & 1.55x & 1.09x & 1.66 & 1.27x \\
    \cmidrule(l{2pt}r{2pt}){2-10}
    & S$^4$C (Ours) & \textbf{2.66x} & \textbf{1.91x} & 2.24x & \textbf{2.12x} & \textbf{2.58x} & \textbf{2.00x} & \textbf{3.86} & \textbf{2.26x} \\
    \midrule
    \multirow{8}*{\textbf{13b}} 
    & EAGLE~\cite{li2024eagle} & 2.55x & 2.08x & \textbf{2.33x} & 1.93x & 2.65x & 2.08x & 3.64 & 2.27x \\
    & Hydra~\cite{ankner2024hydra} & 2.46x & 1.90x & 1.93x & 1.96x & 2.48x & 1.92x & 3.35 & 2.12x \\
    & Medusa~\cite{cai2024medusa} & 1.96x & 1.66x & 1.63x & 1.63x & 2.00x & 1.58x & 2.39 & 1.75x \\
    & PLD~\cite{saxena2023prompt} & 1.47x & 1.02x & 2.19x & 1.03x & 1.57x & 1.71x & 1.68 & 1.48x \\
    & SPS~\cite{leviathan2023fast} & 1.60x & 1.13x & 1.68x & 1.39x & 1.53x & 1.67x & 2.18 & 1.49x \\
    & REST~\cite{he2023rest} & 1.52x & 1.17x & 1.37x & 1.53x & 1.19x & 1.55x & 1.82 & 1.38x \\
    & Lookahead~\cite{fu2024break} & 1.30x & 1.06x & 1.20x & 1.12x & 1.48x & 1.12x & 1.63 & 1.22x \\
    \cmidrule(l{2pt}r{2pt}){2-10}
    & S$^4$C (Ours) & \textbf{2.86x} & \textbf{2.10x} & 2.22x & \textbf{2.11x} & \textbf{3.01x} & \textbf{2.15x} & \textbf{3.98} & \textbf{2.41x} \\
    \midrule
    \multirow{8}*{\textbf{33b}} 
    & EAGLE~\cite{li2024eagle} & 2.80x & 2.08x & 2.43x & 2.19x & 3.00x & 2.22x & 3.38 & 2.46x \\
    & Hydra~\cite{ankner2024hydra} & 2.59x & 2.01x & 2.04x & 2.11x & 2.71x & 2.06x & 3.24 & 2.26x \\
    & Medusa~\cite{cai2024medusa} & 1.98x & 1.73x & 1.64x & 1.66x & 2.07x & 1.62x & 2.33 & 1.79x \\
    & PLD~\cite{saxena2023prompt} & 1.44x & 1.06x & 2.00x & 1.07x & 1.55x & 1.45x & 1.55 & 1.42x \\
    & SPS~\cite{leviathan2023fast} & 1.75x & 1.28x & 1.76x & 1.53x & 1.69x & 1.68x & 2.01 & 1.61x \\
    & REST~\cite{he2023rest} & 1.63x & 1.27x & 1.45x & 1.61x & 1.30x & 1.61x & 2.01 & 1.61x \\
    & Lookahead~\cite{fu2024break} & 1.32x & 1.08x & 1.20x & 1.16x & 1.54x & 1.15x & 1.61 & 1.24x \\
    \cmidrule(l{2pt}r{2pt}){2-10}
    & S$^4$C (Ours) & \textbf{2.82x} & \textbf{2.18x} & \textbf{2.56x} & \textbf{2.33x} & \textbf{3.18x} & \textbf{2.33x} & \textbf{3.67} & \textbf{2.60x} \\
    \bottomrule
  \end{tabular}
  }
  \caption{Speedup ratio and mean accepted tokens of the Vicuna-v1.3 model (sizes 7B, 13B, and 33B) across six tasks.}
  \label{tab:speedup_results}
\end{table*}

Table~\ref{tab:speedup_results} presents the experimental results, demonstrating S$^4$C's superior performance across various tasks and model sizes. Specifically, S$^4$C achieves the highest speedup ratios of \textbf{2.26x}, \textbf{2.41x}, and \textbf{2.60x} for 7B, 13B, and 33B models, respectively, outperforming all baseline methods. 
Moreover, the mean accepted token length with S$^4$C is notably higher across all model sizes (3.86, 3.98, and 3.67 for 7B, 13B, and 33B, respectively), indicating its ability to generate longer, more coherent token sequences. This improvement suggests that S$^4$C enables the target model to accept a greater number of tokens per inference step, thereby enhancing efficiency while maintaining output quality.

Overall, the results validate S$^4$C’s robustness and efficiency in accelerating inference across diverse and complex tasks, highlighting its potential for practical applications in real-world scenarios.

\subsection{Additional Space Consumption (RQ2)}
Dynamic trees have recently gained increasing attention in inference acceleration, with EAGLE2~\cite{li2024eagle2} achieving notable success. However, these improvements come with significant additional memory consumption, often without adequately balancing acceleration gains and resource costs. To address this, we quantified the resource usage of top-performing draft models, including Hydra~\cite{ankner2024hydra}, EAGLE~\cite{li2024eagle}, and EAGLE2~\cite{li2024eagle2}, and compared them using the following efficiency metric:

\begin{equation}
r = \frac{\text{Acceleration\_ratio}}{\text{Extra\_memory}}
\end{equation}

A higher $r$ value indicates greater efficiency, meaning the model achieves a higher acceleration ratio with minimal extra memory, while a lower $r$ implies higher resource demands for comparable acceleration.

\begin{table}[h]
  \centering
  \scalebox{0.9}{
  \begin{tabular}{l|c|c|c}
    \toprule
    Model & Extra Memory  & Acceleration  & Efficiency \\
          & (GB)          & Ratio (×)     & $r$ (↑) \\
    \midrule
    Hydra  & 12.31  & 2.06  & 0.1674 \\
    EAGLE  & 9.30   & 2.12  & 0.2279 \\
    EAGLE2 & 10.54  & \textbf{2.38}  & 0.2258 \\
    S$^4$C (Ours) & \textbf{9.26}   & 2.26  & \textbf{0.2440} \\
    \bottomrule
  \end{tabular}
  }
  \caption{Relationship between extra memory consumption and acceleration.}
  \label{tab:table-Q2}
\end{table}

As shown in Table~\ref{tab:table-Q2}, S$^4$C achieves the highest efficiency with an $r$ value of \textbf{0.2440}, demonstrating its superior balance between acceleration and memory usage. Although EAGLE2 achieves the highest acceleration ratio of 2.38x, it requires an additional 10.54GB of memory, resulting in a lower efficiency score of 0.2258. This suggests that the impact of dynamic trees on acceleration is constrained by their substantial resource requirements.

In conclusion, our findings highlight that while additional memory can enhance acceleration, the efficiency of this trade-off varies. S$^4$C achieves a favorable balance, offering high acceleration with minimal resource overhead, making it a promising approach for optimizing performance and memory efficiency in speculative sampling models.

\subsection{Different Validation Trees (RQ3)}
\label{sec:rq3}
\begin{figure}[htbp]
    \centering  
    \subfigure[Acceleration ratio.]{
    \includegraphics[width=0.45\linewidth]{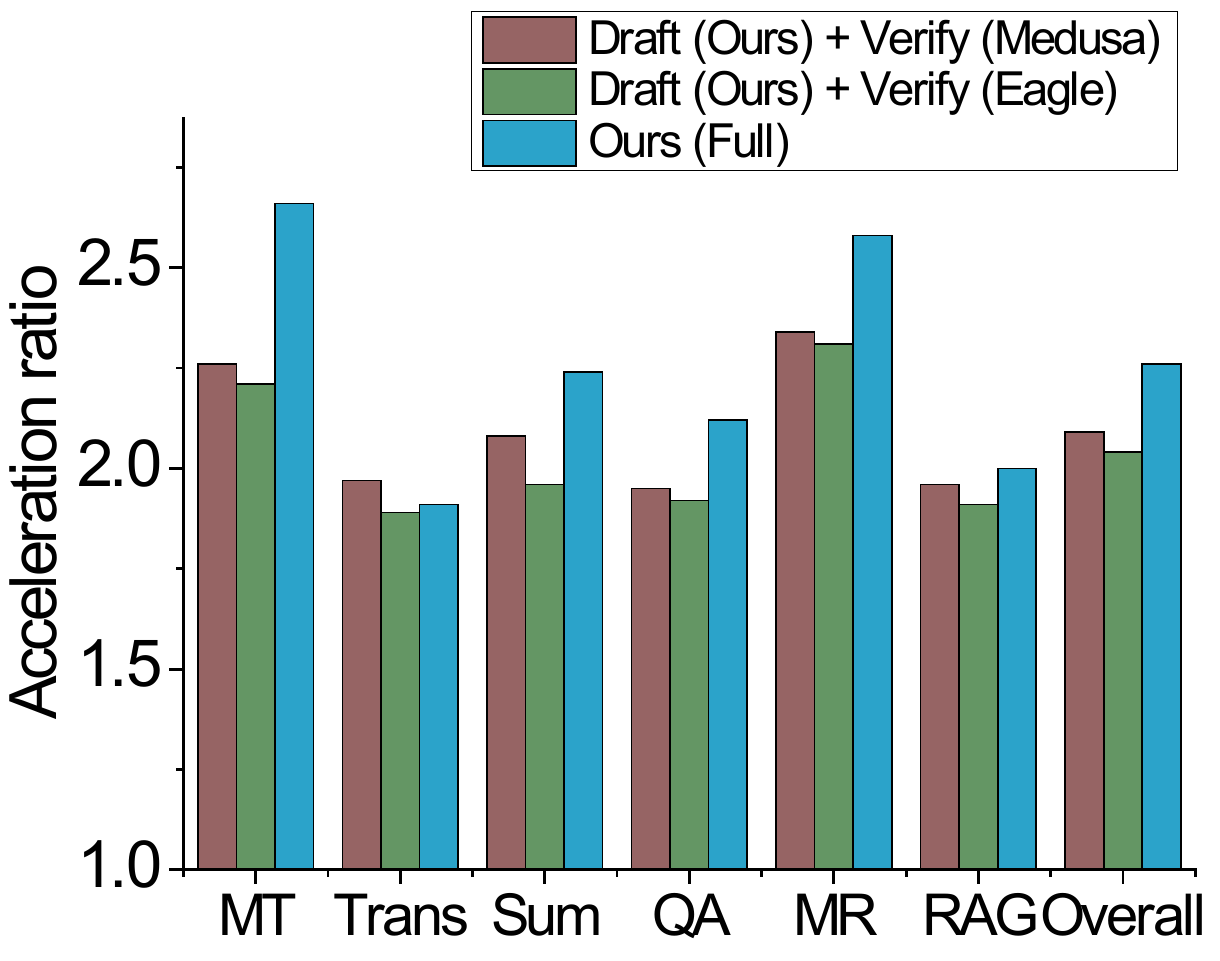}
    \label{fig-Q2-1-1}}
    \subfigure[Accepted length ratio.]{
    \includegraphics[width=0.45\linewidth]{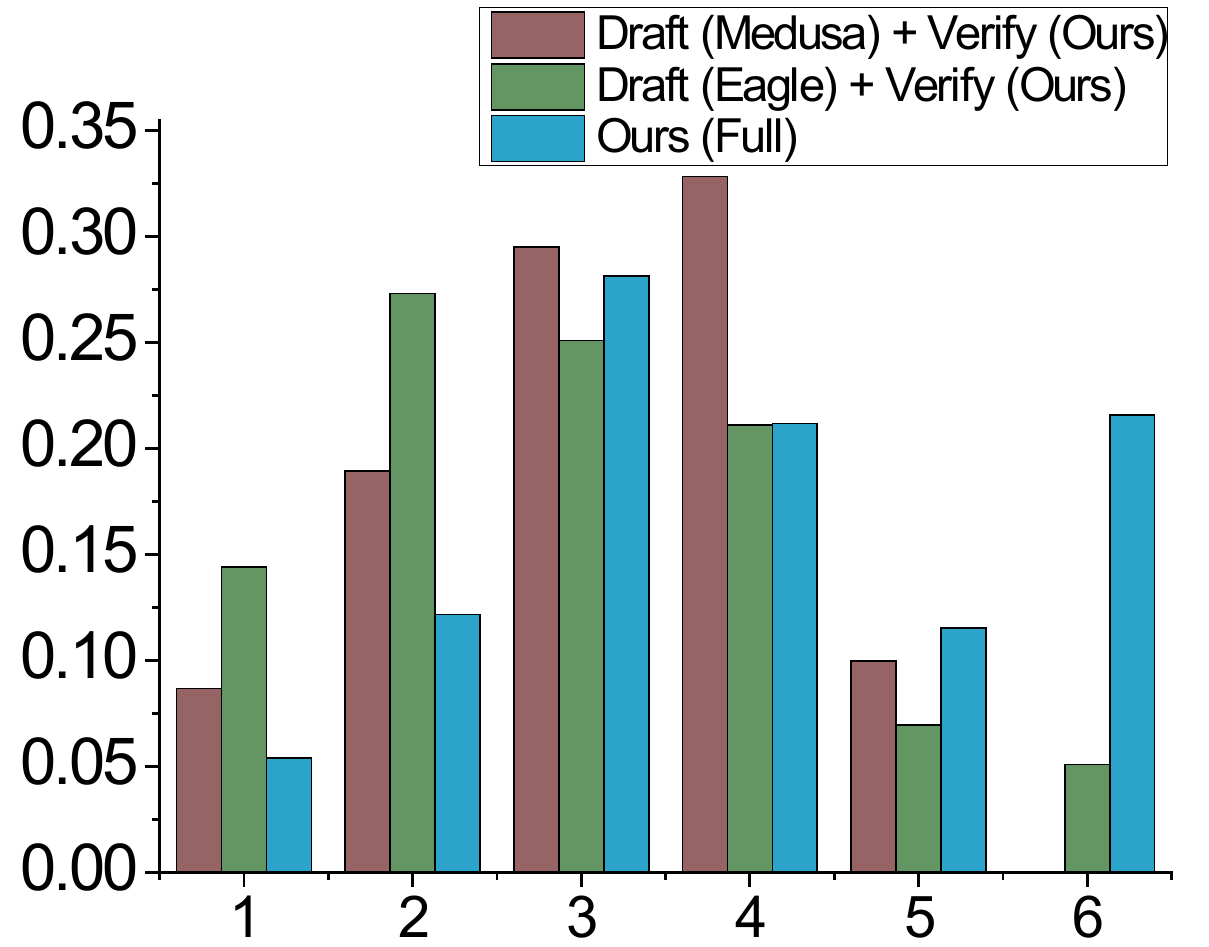}
    \label{fig-Q2-1-2}}
    \caption{Performance comparison of different validation trees.}
    \label{fig-validation-trees}
\end{figure}

In this section, we systematically evaluated the impact of different validation tree structures on the performance and efficiency of preliminary model integration. The experiment is based on the Vicuna-v1.3-7b model, where the draft model is the S$^4$C we trained. We selected Medusa~\cite{cai2024medusa}, Eagle~\cite{li2024eagle}, and our proposed validation tree structure for comparative analysis. The experimental results are shown in Figure \ref{fig-validation-trees}, which clearly demonstrate the performance of different validation tree configurations in terms of acceleration ratio and average acceptance length.

Figure~\ref{fig-Q2-1-1} illustrates the acceleration ratios, a key metric for measuring inference speedup achieved by different validation trees. Our approach demonstrates the highest acceleration ratios, significantly outperforming other methods across the MT, Sum, QA, and MR tasks. 

Figure~\ref{fig-Q2-1-2} presents the distribution of average acceptance lengths, reflecting the number of tokens accepted by the model for each validation tree. 
Compared to other methods, our approach exhibits a higher proportion of acceptance lengths of 5 and 6, while lower proportions for lengths of 1, 2, and 3. This indicates that our method achieves longer average acceptance lengths, contributing to enhanced efficiency and improved generation continuity.

\subsection{Different Draft Models (RQ4)}
To further evaluate the effectiveness of our approach, we integrated our continuous verification tree with various draft models to analyze their impact on overall performance. 
The experimental design is similar to Section \ref{sec:rq3}, based on the Vicuna-v1.3-7b model. The draft models are selected from Medusa, Eagle, and the S$^4$C we trained.
This evaluation aims to determine how different draft model architectures contribute to inference efficiency and token acceptance.

\begin{figure}[!htbp]
\centering  
\subfigure[Acceleration ratio.]{
\includegraphics[width=0.45\linewidth]{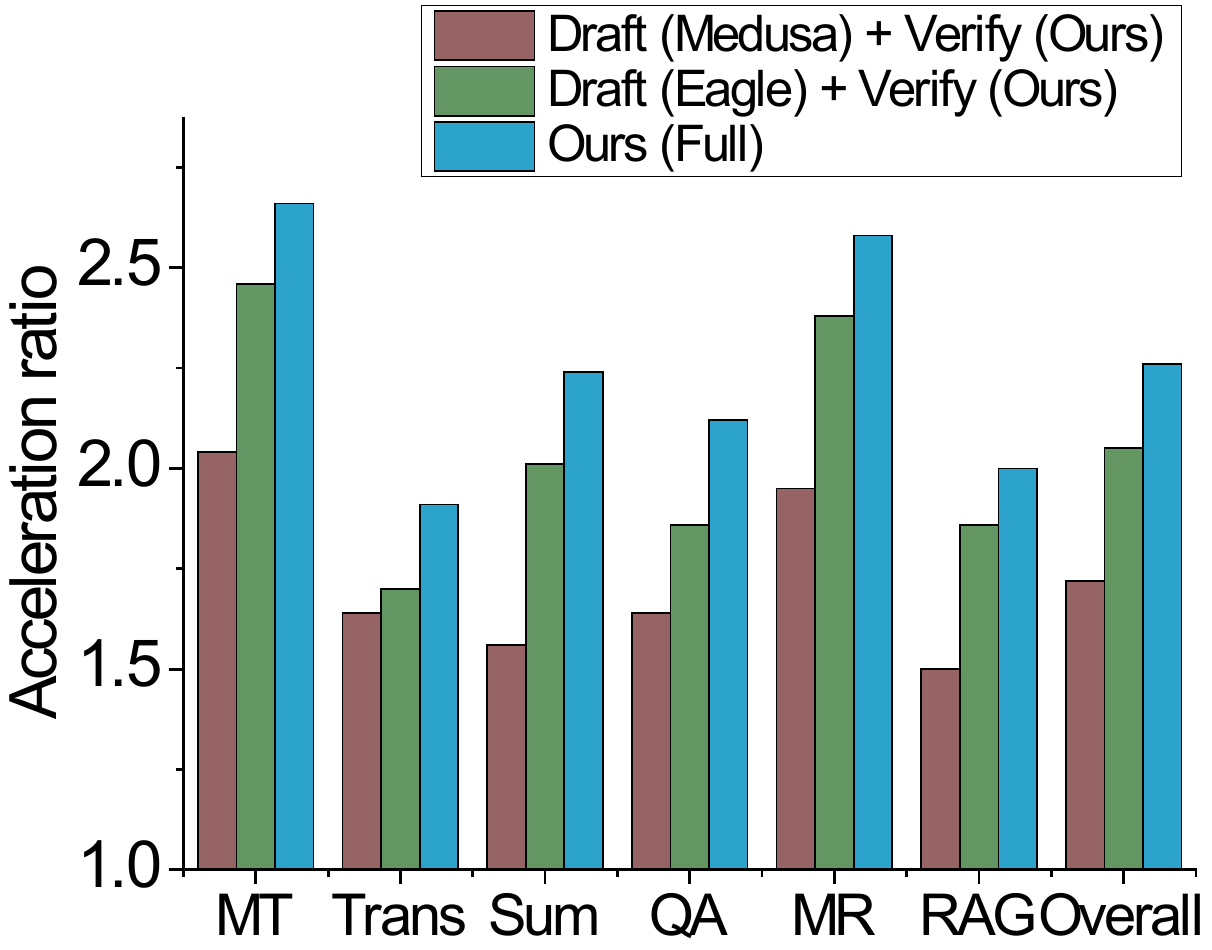}
\label{fig-Q2-2-1}}
\subfigure[Accepted length ratio.]{
\includegraphics[width=0.45\linewidth]{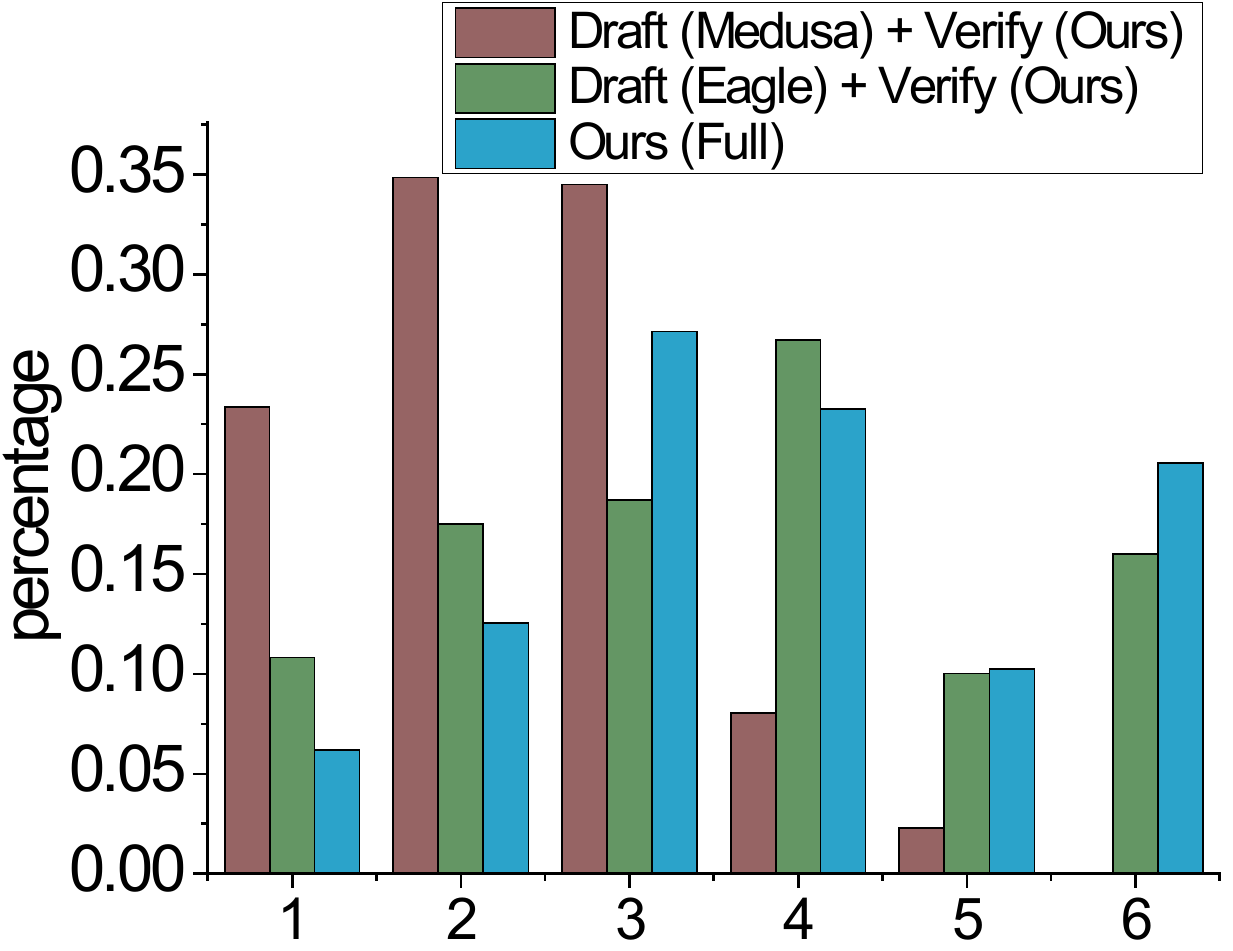}
\label{fig-Q2-2-2}}
\caption{Comparison of different draft models.}
\end{figure}

Figure~\ref{fig-Q2-2-1} presents the acceleration ratios across multiple tasks, including MT, Trans, Sum, QA, MR, and RAG, as well as an overall comparison. Our proposed model consistently achieves the highest acceleration ratio across all tasks, with the most significant improvements observed in MT and MR. Among the baseline models, Eagle ranks second, while Medusa demonstrates the lowest acceleration performance.

Figure~\ref{fig-Q2-2-2} illustrates the distribution of accepted token lengths under the same experimental settings. The majority of accepted tokens for Medusa are concentrated in shorter lengths (1, 2, 3), while Eagle shows a more balanced distribution. In contrast, our model excels in producing longer accepted token sequences, particularly in the third and sixth categories. Overall, our approach achieves the longest average acceptance length, indicating improved coherence and efficiency in token generation.

\subsection{Capabilities (RQ5)}
\begin{figure}[!htbp]
\centering  
\subfigure[Acceleration ratio.]{
\includegraphics[width=0.45\linewidth]{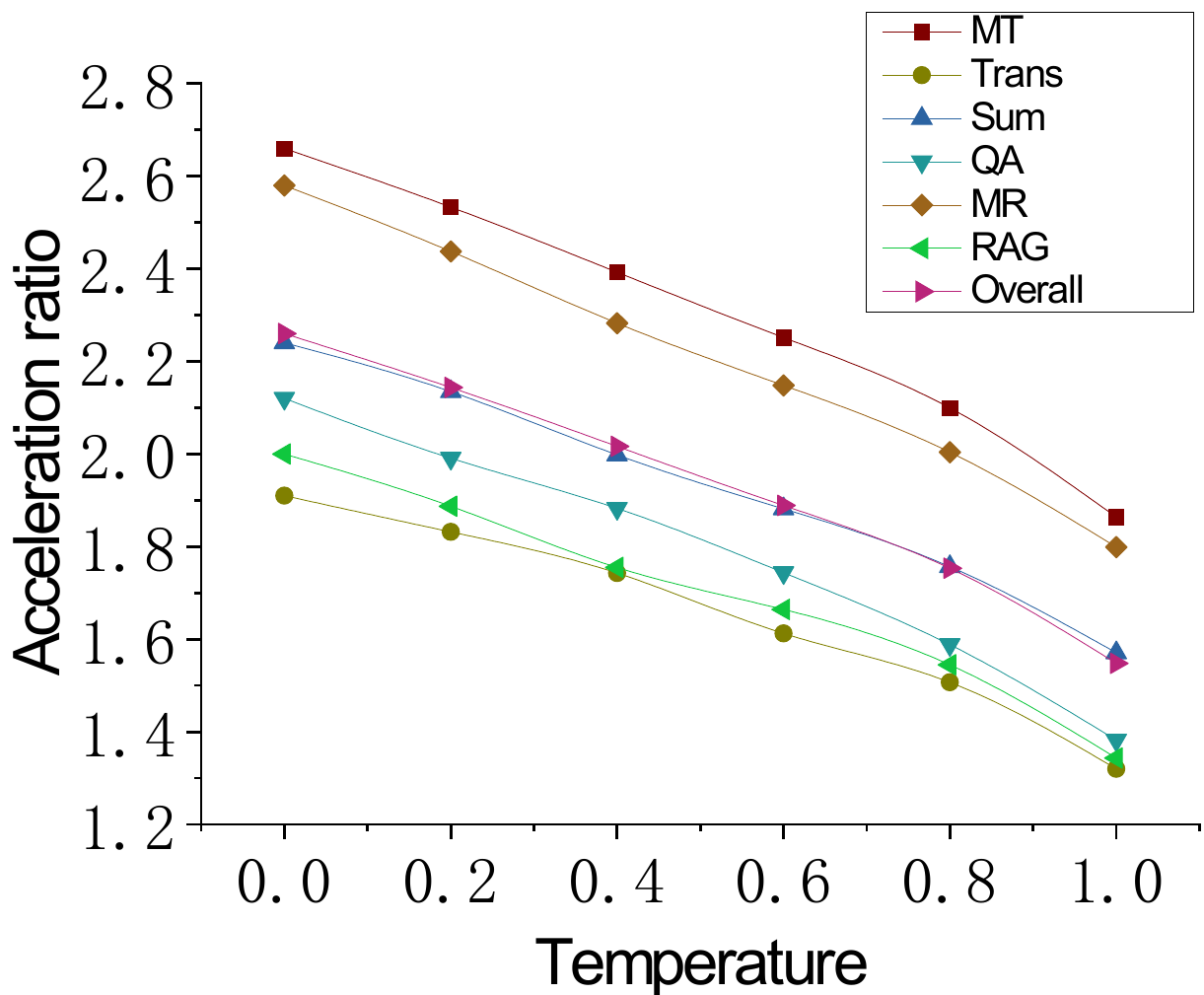}
\label{fig-Q5-1}}
\subfigure[Mean accepted tokens.]{
\includegraphics[width=0.45\linewidth]{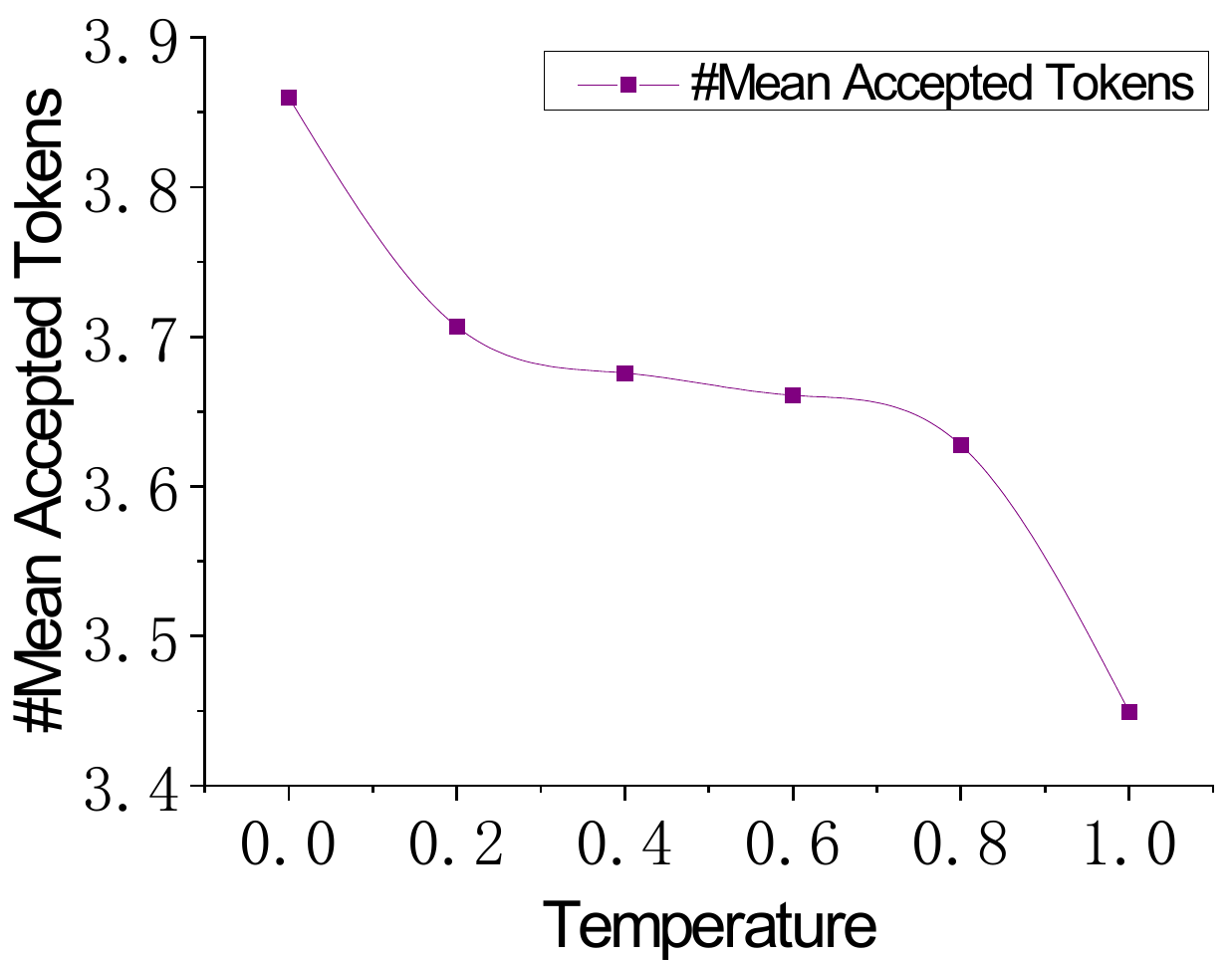}
\label{fig-Q5-2}}
\caption{S$^4$C performance under different temperature settings.}
\label{fig-Q5}
\end{figure}

Different token selection strategies can be employed during the inference phase of large language models, including selecting the most probable token, beam search, and sampling methods. Inference sampling, in particular, can be influenced by temperature settings, which control the randomness of token selection. In this section, we evaluate the performance of S$^4$C under varying temperature settings, as shown in Figure~\ref{fig-Q5}. 

Our analysis reveals a clear trend: as the temperature increases, the performance of S$^4$C gradually declines (Figure~\ref{fig-Q5-1}). This degradation is attributed to the increased randomness in token selection, which reduces the number of tokens accepted by the target model (Figure~\ref{fig-Q5-2}). As a result, the acceleration ratio decreases due to a lower number of retained tokens, highlighting the sensitivity of S$^4$C to temperature variations.

\subsection{Case Study (RQ6)}
\begin{figure}[!htbp]       
	\centering
    \includegraphics[width=0.45\textwidth]{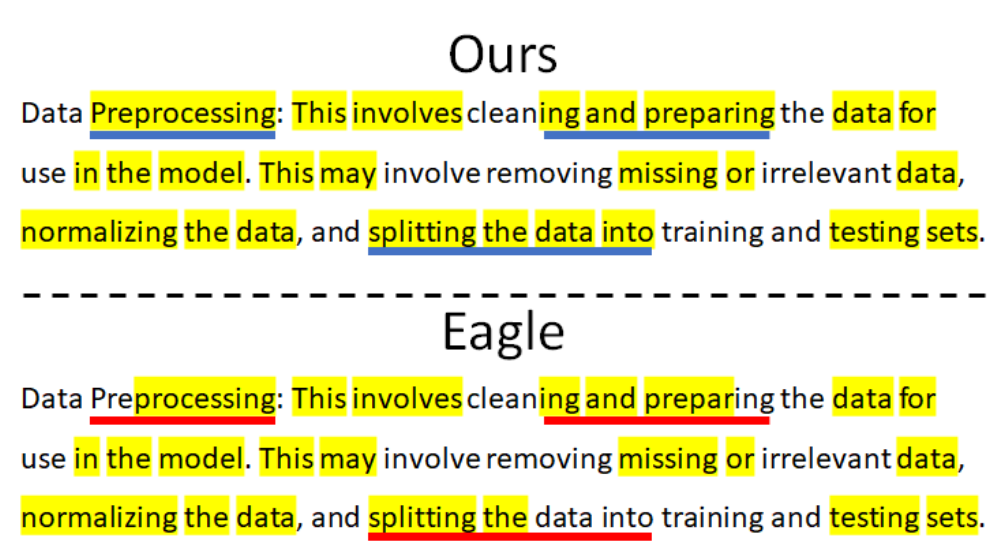} 
	\caption{Case study illustration. Highlighted words show accepted tokens, and underlined words show differences.}  
	\label{fig-Q6}
\end{figure}

In this section, we evaluate the performance of our method by examining the draft tokens accepted by the model, comparing it with previous approaches, and validating the motivation behind enhancing both syntactic and semantic coherence. The experiment specifically highlights token continuity, syntactic, and semantic coherence under different conditions. The results are illustrated in Figure~\ref{fig-Q6}.

Our method successfully generates continuous token sequences, such as ``splitting the data into'', which maintains both syntactic and semantic coherence by ensuring logical consistency with the preceding context. In contrast, Eagle~\cite{li2024eagle} struggles with token continuity, often accepting only partial sequences like ``pre-processing'' and ``prepar-ing''.
This comparison highlights a key limitation of Eagle~\cite{li2024eagle}: despite focusing on semantic coherence through an autoregressive approach, it fails to preserve syntactic continuity, resulting in fragmented token acceptance. In contrast, our method effectively balances both syntactic and semantic coherence, leading to a higher acceptance rate of draft tokens, longer accepted token sequences, and an overall improvement in speed. This highlights the importance of considering both aspects to overcome the limitations of previous approaches that prioritize only one dimension.
\section{Related Work}
\label{sec:related_work}

\subsection{Speculative Sampling}
The draft-then-verify decoding strategy was first introduced by Stern et al.~\cite{stern2018blockwise}, while speculative sampling~\cite{xia2023speculative,leviathan2023fast,chen2023accelerating} extended this concept to non-greedy sampling, ensuring the preservation of the original output distribution. Xia et al.~\cite{xia2024unlocking} provided a comprehensive survey of recent advancements in speculative sampling, categorizing the drafting process into two main approaches: \textit{Independent Drafting} and \textit{Self-Drafting}.
In the context of Independent Drafting, SpecDec~\cite{xia2023speculative} pioneered the use of independent models for drafting, striking a balance between accuracy and efficiency. Leviathan et al.~\cite{leviathan2023fast} demonstrated the acceleration of T5-XXL inference by employing T5-small as a drafting model. 
These methods use a lightweight, pre-trained LLM that does not require additional training or modification, facilitating the seamless adoption of speculative decoding in various applications\cite{leviathan2023fast,DBLP:journals/corr/abs-2308-04623,DBLP:conf/nips/SunSRBJY23,chen2023accelerating}.

In summary, speculative sampling leverages independent models for draft generation, improving inference efficiency while maintaining the accuracy of the output distribution.  
These approaches~\cite{xia2023speculative,leviathan2023fast,chen2023accelerating} balance efficiency and accuracy, making speculative decoding widely applicable across various tasks.
\subsection{Self-Drafting}
Our research primarily falls under the Self-Drafting category, which focuses on utilizing the target LLM itself for efficient drafting~\cite{xia2024unlocking}. Blockwise Decoding~\cite{stern2018blockwise} and Medusa~\cite{cai2024medusa} introduced Feed-Forward Network (FFN) heads within the Transformer decoder, enabling parallel token generation at each decoding step. However, this parallel structure often results in suboptimal draft quality. To address this issue, Hydra~\cite{ankner2024hydra} improves Medusa by enhancing the correlation between draft head predictions, thereby increasing draft accuracy.
In contrast, Eagle~\cite{li2024eagle} reintroduces an autoregressive structure while leveraging the dual characteristics of tokens and functions, enhancing drafting precision. Furthermore, Eagle2~\cite{li2024eagle2} introduces a confidence-based dynamic tree mechanism to optimize the acceptance length, improving the overall efficiency of speculative sampling.

In summary, recent advancements in the Self-Drafting category have shown a clear trend towards improving the efficiency and quality of draft generation through innovative decoding mechanisms. These developments ~\cite{DBLP:journals/tmlr/YangLCP024,DBLP:conf/acl/Zhang00S0CM24,DBLP:journals/corr/abs-2310-12072,DBLP:conf/acl/SantilliSPMMMR23,DBLP:journals/corr/abs-2311-13581} highlight the ongoing efforts to balance parallelism and sequential dependencies in draft generation, aiming to achieve both high efficiency and high quality in the context of LLM-based drafting.

\section{Conclusion and Future Work}
\label{sec:conclusion}
This paper presents S$^4$C, an efficient conjectural sampling framework that employs a multi-head autoregressive model structure to model both syntactic and semantic coherence for rapid token generation. 
S$^4$C preserves the original output distribution of the target LLM while significantly accelerating the generation process.
Experimental results on the Spec-bench benchmark show that S$^4$C outperforms state-of-the-art methods, achieving an acceleration ratio of 2.26x to 2.60x over standard autoregressive decoding.
For future work, we aim to enhance S$^4$C’s adaptability to diverse LLM architectures and optimize the trade-off between draft quality and verification complexity. Further research will investigate S$^4$C’s applicability to multimodal tasks.

\bibliography{custom}

\appendix



\end{document}